\documentclass[letterpaper, 10 pt, conference]{ieeeconf}

\IEEEoverridecommandlockouts
\usepackage{cite}
\usepackage{amsmath,amssymb,amsfonts}
\usepackage{algorithmic}
\usepackage{graphicx}
\usepackage{textcomp}
\usepackage[table]{xcolor}
\usepackage{comment}
\usepackage[ruled,vlined]{algorithm2e}
\usepackage{caption}
\usepackage{array}
\usepackage{subcaption}
\usepackage{arydshln}

\usepackage{multirow}

\usepackage{tensor}
\newcommand{\eg}{\emph{e.g.},}
\newcommand{\ie}{\emph{i.e.},}

\usepackage[bookmarks=false]{hyperref}
\usepackage{float}
\usepackage{rotating}
\usepackage{nicematrix}  %
\usepackage{mathtools}
\usepackage{amsmath}
\usepackage{pifont}
\usepackage{cuted}
\usepackage{dblfloatfix}
\usepackage{balance}
\usepackage{titlesec}

\usepackage{dsfont}
\usepackage{mathtools}
\usepackage{booktabs}
\usepackage{graphics}
\usepackage[mathscr]{euscript}

\renewcommand{\vec}[1]{{\mathbf #1}}

\newcommand{\cmark}{\ding{51}}%
\newcommand{\xmark}{\ding{55}}%

\newcommand{\algstep}[1]{\item[]\medskip\hrule\kern 2pt\hbox to \textwidth{\hspace{\labelsep}\textbf{#1}\hfill}\hrule}

\usepackage{xspace}
\makeatletter
\DeclareRobustCommand\bmvaOneDot{\futurelet\@let@token\bmv@onedotaux}
\def\bmv@onedotaux{\ifx\@let@token.\else.\null\fi\xspace}
\makeatother

\definecolor{LightCyan}{rgb}{0.88,1,0.88}
\definecolor{linear_color}{RGB}{220,223,240}
\definecolor{gray_bbox_color}{RGB}{243,243,244}
\newcommand{\coolname}{\textit{M2Distill}}
\definecolor{rebuttal}{rgb}{0,0,1}

\def\eqref#1{Eq.~(\ref{#1})}

\begin{document}

\title{\LARGE \bf M2Distill: Multi-Modal Distillation for Lifelong Imitation Learning}

\author{Kaushik Roy$^{1}$, Akila Dissanayake$^{1,2}$, Brendan Tidd$^{1}$, Peyman Moghadam$^{1,2}$
\thanks{$*$ This work was partially funded by CSIRO's Data61 Science Digital. Authors acknowledge continued support from the CSIRO's Data61 Embodied AI Cluster.}
\thanks{$^1$ CSIRO Robotics, DATA61, CSIRO, Australia. 
E-mails: {\tt\footnotesize \emph{firstname.lastname}@csiro.au}}
\thanks{$^2$ Queensland University of Technology (QUT), Brisbane, Australia. 
E-mails: {\tt\footnotesize \emph{firstname.lastname}@qut.edu.au}}
}

\maketitle

\begin{abstract}

Lifelong imitation learning for manipulation tasks poses significant challenges due to distribution shifts that occur in incremental learning steps. Existing methods often rely on unsupervised skill discovery to construct an ever-growing skill library or distillation from multiple policies, which can lead to scalability issues as diverse manipulation tasks are continually introduced and may fail to ensure a consistent latent space throughout the learning process, leading to catastrophic forgetting of previously learned skills. In this paper, we introduce \coolname{}, a multi-modal distillation-based method for lifelong imitation learning focusing on preserving consistent latent space across vision, language, and action distributions throughout the learning process. By regulating the shifts in latent representations across different modalities from previous to current steps, and reducing discrepancies in Gaussian Mixture Model (GMM) policies between consecutive learning steps, we ensure that the learned policy retains its ability to perform previously learned tasks while seamlessly integrating new skills. Evaluations on the LIBERO lifelong imitation learning benchmark suites, including LIBERO-OBJECT, LIBERO-GOAL, and LIBERO-SPATIAL, demonstrate that our method consistently outperforms prior state-of-the-art methods across all evaluated metrics.

\end{abstract}

\section{Introduction}
\label{sec:intro}

\begin{figure}[!t]
\vspace{-0.5em}
\hspace{-.75em}
\centering
\includegraphics[width=.5\textwidth, scale=1.5]{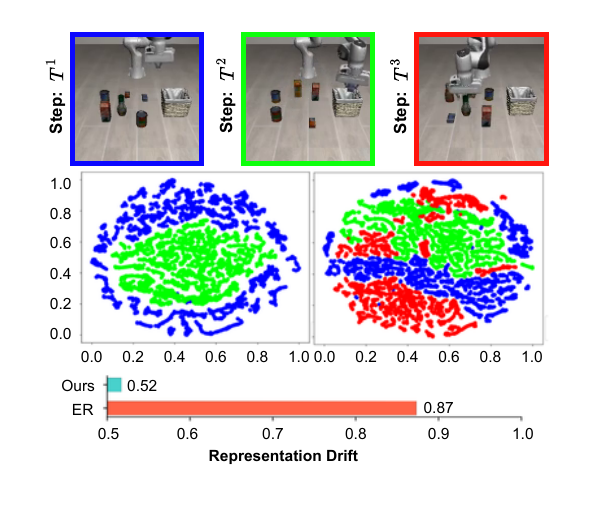}
\vspace{-2.5em}
\caption{t-SNE visualization of latent space deformation for AgentView images using Experience Replay (ER) method across two consecutive steps (\ie{} $T^2$ and $T^3$) in a lifelong imitation learning scenario on LIBERO-OBJECT. The t-SNE plots highlight significant shifts in latent representations, contributing to catastrophic forgetting, while the accompanying bar plot shows reduced representation drift in \coolname{} during the sequential learning of manipulation tasks. %
} 
\label{fig:fspace_deform} 
\vspace{-1.0em}
\end{figure}

In recent years, the field of robotics has made significant strides in creating intelligent systems capable of performing complex tasks autonomously. Among these advancements, Imitation Learning (IL) has become a popular and effective paradigm for robots to learn complex behaviors by observing and mimicking human demonstrations~\cite{stepputtis2020language,jang2022bc,xie2024decomposing}. 
Recent research further enhances the generalization ability of these methods by leveraging multi-modal inputs, such as vision, language, and actions, and the recent large vision-language-action (VLA) models~\cite{zitkovich2023rt, team2024octo}. 

Despite the impressive performance of IL models, the current state-of-the-art IL models focus on either learning from a single task or a known set of tasks in advance. This impedes their applicability in complex real-world settings, where robots need to continually learn new tasks as they arrive while retaining the models’ performance on previously learned tasks. This is known as Lifelong Imitation Learning (LIL). Recently, a few studies~\cite{gao2021cril, zhu2022bottom, liu2024libero, wan2024lotus} show promising results in addressing imitating learning from sequentially arriving tasks while avoiding catastrophic forgetting. 
Catastrophic forgetting refers to undesirable behavior of the neural networks in which newly acquired skills (\ie{} knowledge) can degrade the preservation of previous ones \cite{french1999catastrophic, kirkpatrick2017overcoming}. 

Recent LIL methods, such as ER~\cite{chaudhry2019tiny}, method replay a subset of examples from previous tasks alongside new task data. However, imbalanced data distribution can cause the latent representations of past tasks to drift, leading the trained policy to favor the current task and reducing its performance on earlier tasks. Methods like BUDS~\cite{zhu2022bottom}, and LOTUS~\cite{wan2024lotus} rely on unsupervised skill discovery and integration, but maintaining an ever-expanding skill library becomes computationally expensive over time. PolyTask~\cite{haldar2023polytask} addresses this by distilling skills from task-specific policies, but this approach requires access to all previously learned policies, making it resource-intensive and unrealistic. 

To investigate the factors contributing to catastrophic forgetting in these scenarios, we visualize t-SNE plots of latent representations for AgentView using a ResNet18 backbone from ER policies at steps 2 and 3, respectively. The results, shown in Figure~\ref{fig:fspace_deform}, reveal significant deformation in the latent space, with representations of prior tasks drifting considerably. This trend is consistent across different modalities and motivates us to design a multi-modal distillation framework to preserve both latent representations and action distributions as we continuously train our policy on new manipulation tasks.

To address these challenges, we introduce \coolname{}, a multi-modal distillation method for lifelong imitation learning. 
The primary objective of the proposed multi-modal distillation method is to learn a consistent latent space across multiple modalities (language, vision, and joints) that can enable robotic systems to continuously learn new manipulation skills while effectively retaining acquired knowledge from previous tasks. 
The distillation and alignment of knowledge across modalities are achieved by minimizing the Euclidean distance between feature embeddings extracted from the old and current models. %
Additionally, ensuring action consistency between the old and current policy is crucial for smooth learning and adaptation without forgetting. Our method addresses this by minimizing the Kullback-Leibler (KL) divergence between the Gaussian Mixture Model (GMM) policy of the old and current models. This approach ensures that the actions predicted by both models remain closely aligned for previously learned tasks, maintaining consistent performance on these tasks while accommodating new ones.

\noindent
Overall, our contributions in this paper are as follows:
\begin{itemize}
  \item We present \coolname{}, a lifelong imitation learning framework that incorporates a multi-modal feature and action distillation strategy. This framework preserves the consistency of the latent space (language, vision and joints) and action distributions of the GMM policies while learning a series of manipulation tasks from human demonstrations in memory-constrained settings.
  \item Our proposed method demonstrates significant performance improvements across three LIBERO lifelong imitation learning benchmark suites, such as LIBERO-OBJECT, LIBERO-GOAL and LIBERO-SPATIAL. %
\end{itemize}

\label{sec:rel_work}
\section{Related Work}

Imitation Learning (IL), also known as Learning from Human Demonstration, is a machine learning paradigm where a robot learns to perform tasks by mimicking the actions of an expert demonstrator~\cite{hussein2017imitation, schaal1996learning, billard2016learning}. The primary goal of imitation learning is to learn a policy that maps observations to actions and replicates the expert's actions. Lifelong Imitation Learning (LIL) extends this concept by focusing on the continuous acquisition of skills over time while retaining previously learned knowledge. In LIL, robots are designed to adapt to new tasks and environments without forgetting earlier skills, addressing the issue of catastrophic forgetting~\cite{kirkpatrick2017overcoming, french1999catastrophic}.

Catastrophic forgetting is well-studied for various problems in computer vision, including classification~\cite{roy2023cl3}, detection~\cite{shmelkov2017incremental}, and semantic segmentation~\cite{roy2023subspace}, within lifelong learning scenarios. In the literature of lifelong learning, dynamic architecture~\cite{yoon2017lifelong, douillard2022dytox}, regularization~\cite{ahn2019uncertainty, roy2023l3dmc}, and memory-replay~\cite{rebuffi2017icarl, rolnick2019experience, kirkpatrick2017overcoming, gao2021cril} based approaches have been proposed to tackle the catastrophic forgetting. Regularization based methods control the changes in the network's weight by introducing new regularization terms. Memory-replay based strategies store a subset of past examples and replay with new examples.

Lifelong learning has shown promise in the field of robotics; however, the volume of research specifically focusing on lifelong imitation learning remains limited. 
ER~\cite{chaudhry2019tiny} preserves a limited number of past trajectories and replays them in conjunction with new trajectories from the ongoing manipulation task. In contrast, CRIL~\cite{gao2021cril} leverages generative adversarial networks (GAN~\cite{lesort2019generative}) to generate the first frame of each trajectory and relies on an action-conditioned video prediction network to predict future frames using states and actions from the deep generative replay (DGR~\cite{shin2017continual}) policy.
BUDS~\cite{zhu2022bottom} introduces a technique for skill discovery in robot manipulation tasks that do not require pre-segmented demonstrations. By employing a bottom-up approach, it autonomously identifies and organizes skills from unsegmented, long-horizon demonstrations, enabling robots to effectively handle complex and extended manipulation tasks.
LOTUS~\cite{wan2024lotus} allows robots to continuously learn and adapt to new tasks by leveraging unsupervised skill discovery and integration. It utilizes an open-vocabulary vision model for skill discovery and a meta-controller for skill integration.
PolyTask~\cite{haldar2023polytask} proposes a method for learning unified policies across multiple tasks through behavior distillation. This approach distills knowledge from expert policies into a single policy, enabling the robot to efficiently perform various tasks with a generalized model.
However, these LIL approaches still need to tackle challenges related to scalability and the effective integration of new skills across varied environments.

\vspace{-1mm}
\section{Problem Formulation} 
\vspace{-0.4mm}
The Lifelong Robot Learning problem extends the traditional robot learning framework by requiring a robot to continuously acquire, adapt, and retain knowledge across a sequence of tasks \( \{T^1, \dots, T^K\} \) 
over its operational lifespan. This robot learning problem is formulated as a finite-horizon Markov Decision Process (MDP), denoted by \( \mathcal{M} = (\mathcal{S}, \mathcal{A}, \mathcal{T}, H, \mu_0, R) \), where \( \mathcal{S} \) represents the state space, \( \mathcal{A} \) is the action space, \( \mathcal{T} : \mathcal{S} \times \mathcal{A} \rightarrow \mathcal{S} \) is the transition function, \( H \) is the maximum horizon for each episode of a task, \( \mu_0 \) is the initial state distribution, and \( R : \mathcal{S} \times \mathcal{A} \rightarrow R \) is the reward function. Given that the reward function \( R \) is often sparse, a goal predicate \( g: \mathcal{S} \rightarrow \{0, 1\} \) is used to indicate whether a goal has been achieved.
\begin{figure*}[!t]
\hspace{-1em}
\centering
\includegraphics[width=\textwidth, scale=1]{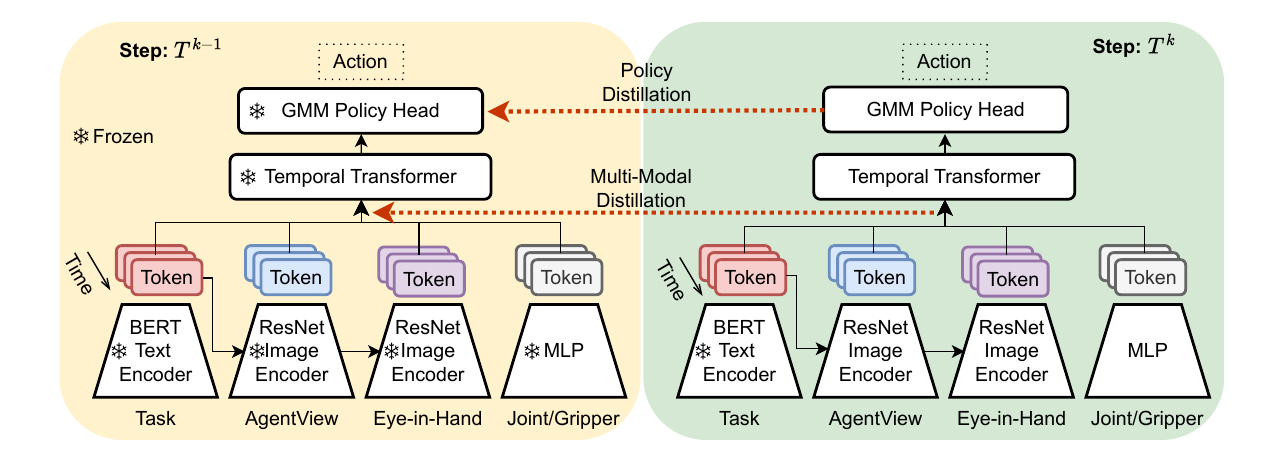}
\caption{Overview of our proposed \coolname{} method. %
Multi-modal distillation aligns the latent representations from different input modality encoders (\eg{} Task, AgentView, Eye-in-Hand, Joint, and Gripper information), while policy distillation maps the action distribution of the GMM policy between incremental steps $T^{k-1}$ and $T^{k}$.
} 
\label{fig:mmd_framework} 
\vspace{-1em}
\end{figure*}
In the context of lifelong learning~\cite{kirkpatrick2017overcoming}, the robot must develop a single policy \( \pi \) that can adapt to the specific requirements of each task \( T^k \). This policy is conditioned on the task at hand, allowing the robot to tailor its policy to meet the unique objectives of each task while remaining consistent in its structure. Each task \( T^k = (\mu_0^k, g^k) \) is characterized by its own initial state distribution \( \mu_0^k \) and goal predicate \( g^k \), while the state space \( \mathcal{S} \), action space \( \mathcal{A} \), transition function \( \mathcal{T} \), and time horizon \( H \) remain unchanged across all tasks. The robot’s broader objective is to maximize its performance across all tasks, which can be mathematically represented as:
\begin{align}
\max_{\pi} J_{}(\pi) = \frac{1}{K} \sum_{k=1}^{K} \mathbb{E}_{s_t^k, a_t^k \sim \pi(\cdot; T^k), \mu_0^k} \left[ \sum_{t=1}^{L^k} g^k(s_t^k) \right],
\end{align}
where $L^k$ is the length of the trajectory. $s_t^k, \text{and}~a_t^k$ are the state and action pair sampled from policy $\pi$ conditioned with task $T^k$. %

\noindent
\textbf{Lifelong Imitation Learning.} In this setting, a robot sequentially trains a policy $\pi$ through imitation learning over multiple tasks. %
For each task \( T^k \), the robot receives a small dataset of \( N \) expert demonstrations, denoted by \( D^k = \{\tau^k_i\}_{i=1}^{N} \), along with a corresponding language description \( l^k \). Expert demonstrations are collected via teleoperation, with each trajectory \( \tau^k_i \) consisting of state-action pairs \( \{(s_t, a_t)\}_{t=1}^{L^k} \), where \( L^k < H \). The policy is trained using a behavioral cloning loss~\cite{bain1995framework}, which aims to mimic the actions demonstrated in the dataset.
\begin{align}
    \hspace{-.75em}
    \min_{\pi} J(\pi) = \frac{1}{K} \sum_{k=1}^{K} \mathbb{E}_{s_t, a_t \sim D^k} \left[ \sum_{t=1}^{L^k} -\log \pi(a_t | s_{\leq t}; T^k) \right].
    \hspace{-.75em}
\end{align}

One of the challenges in lifelong imitation learning is that, as the robot progresses to new tasks, it loses direct access to the previous tasks \( T^1, \dots, T^{k-1} \)~\cite{wang2024comprehensive}. This limitation necessitates that the robot not only performs well on the current task but also effectively retains and transfers knowledge from prior tasks, ensuring that the policy evolves to support future learning. The goal is to balance leveraging previously acquired knowledge and adapting to new challenges, thereby optimizing the robot’s overall learning efficiency and performance over its entire operational lifespan.

\section{Proposed Method - \coolname{}}

In our lifelong imitation learning framework, we prioritize maintaining a consistent low-dimensional latent space across different modalities to address drift in the latent representation distributions. We hypothesize that distilling latent representations from prior models using expert demonstrations can help mitigate shifts in data distributions during incremental learning. Furthermore, we aim to preserve the action distribution from previously learned manipulation tasks while acquiring new skills, ensuring that the knowledge gained from expert demonstrations is retained and leveraged throughout the learning process.

Our architecture for learning manipulation tasks from demonstrations is based on the ResNet-T design from \cite{liu2024libero}, incorporating task embeddings from a pre-trained language transformer, two image encoders for different camera streams, and additional modality encoders for joint positions and gripper state data. The multi-modal tokens from various time steps are processed by a temporal transformer, with the final token passed to a GMM policy head to produce an action distribution. In our approach, we focus on retaining both the multi-modal token distributions and the GMM policy head’s action distribution throughout incremental learning. Figure~\ref{fig:mmd_framework} depicts the overall architecture of our proposed distillation-based LIL approach. The following discussion covers multi-modal distillation, after which we present policy distillation.

\subsection{Multi-modal Distillation}
During the distillation process, we pass the same RGB images through the old and new ResNet18~\cite{he2016deep} encoders, and our proposed distillation strategy is applied to the resulting latent representations. We minimize the squared $L_2$ norm of the difference between the old and new latent representations, ensuring that the latent space for expert demonstrations in previously learned tasks remains intact. This constraint preserves performance on prior tasks while enabling the acquisition of new skills.

Let’s assume that for an input image batch of size $B$ with $L$ timestamps, the extracted feature vectors from the image encoder of our policy $\pi$ have dimensions $B \times L \times 64$. This holds true for both the current step $k$ and the previous step $k-1$.

For image inputs $\vec{f}^{k}$ and $\vec{f}^{k-1}$ at steps $k$ and $k-1$, and considering both the AgentView and HandEye image views, we have the following loss:

\begin{equation}
\mathcal{L}_{\text{image}} = \mathcal{L}_{\text{AgentView}} + \mathcal{L}_{\text{HandEye}},
\end{equation}
where the loss for each modality $\epsilon \in \{\text{AgentView}, \text{HandEye}\}$ is defined as:
\begin{equation}
\mathcal{L}_{\epsilon \in \{\text{AgentView}, \text{HandEye}\}} = \frac{1}{\text{NL}} \sum_{i=1}^{\text{N}}  \sum_{j=1}^{\text{L}} \| \vec{f}^{k, \epsilon}_{i, j} - \vec{f}^{k-1, \epsilon}_{i, j} \|_2^2 .
\end{equation}

Similarly, we feed the text instructions into a pre-trained BERT~\cite{devlin2018bert} text encoder and project the output into a $64$-dimensional latent space using an MLP. Let $\vec{g}^{k}$ and $\vec{g}^{k-1}$ represent the latent representations of the text instruction at steps $k$ and $k-1$. The distillation loss for the text modality is then:
\begin{equation}
\mathcal{L}_{\text{text}} = \frac{1}{\text{NL}} \sum_{i=1}^{\text{N}} \sum_{j=1}^{\text{L}} \| \vec{g}^{k}_{i,j} - \vec{g}^{k-1}_{i,j} \|_2^2 .
\end{equation}
 
Moreover, we condition our policy on additional modalities (\eg{} joint information and gripper state) along with the image and text. Consequently, we define the following distillation loss to maintain a consistent latent space for this extra modality as well.
\begin{equation}
\mathcal{L}_{\text{extra}} = \sum_{\epsilon \in \{\text{joint}, \text{gripper}\}} \frac{1}{\text{NL}} \sum_{i=1}^{\text{N}} \sum_{j=1}^{\text{L}} \| \vec{h}^{k, \epsilon}_{i,j} - \vec{h}^{k-1, \epsilon}_{i,j} \|_2^2  ,
\end{equation}
where $\vec{h}^{k}$ and $\vec{h}^{k-1}$ represent the latent representations of the given joint and gripper modalities encoded using respective encoder at incremental steps $k$ and $k-1$.

\subsection{Policy Distillation}

In this paper, we prioritize preserving a consistent action distribution for previously learned manipulation tasks throughout the continual learning process. %
We address this by replicating the action distribution of the previous GMM policy within the current GMM policy, which helps maintain consistency in the distribution of action space between the two steps.
This strategy is vital in preventing catastrophic forgetting, where new tasks could potentially disrupt the performance of previously learnt ones. 
By utilizing a Kullback-Leibler (KL) divergence loss between the old model's policy and that of the current model, we can ensure that the predicted actions for past tasks remain aligned with their original action distributions. 

Let $\pi^{k}$ and $\pi^{k-1}$ denote the action distributions of the policy at incremental steps $k$ and $k-1$, respectively. The KL divergence between $\pi^{k}$ and $\pi^{k-1}$ can be formulated as follows:
\begin{align}
\mathcal{L}_{\text{policy}} &= \mathcal{L}_{\text{KL}}(\pi^{k} \| \pi^{k-1}) \\ \notag
&= \mathbb{E}_{a \sim \pi^{k}}\left[ \log \pi^{k}(\vec{a}) - \log \pi^{k-1}(\vec{a}) \right] \\ \notag
& = \int \pi^{k}(\vec{a}) \left[\log \pi^{k}(\vec{a}) - \log \pi^{k-1}(\vec{a}) \right] da .
\end{align}

When $\pi^{k}$ and $\pi^{k-1}$ are Gaussian distributions, this KL divergence has a closed-form solution. For mixtures of Gaussians (GMMs), which is the case in this work, the KL divergence lacks a closed-form expression due to the complexity of the mixture components \cite{hershey2007approximating}. To tackle this issue, we employ Monte Carlo sampling to approximate the KL divergence. Specifically, we draw a set of samples $\{a^{s}\}_{s=1}^N$ from the distribution $\pi^{k}$, and estimate the KL divergence by averaging the log difference between $\pi^{k}(a)$ and $\pi^{k-1}(a)$ over these samples as follows.
\begin{align}
\mathcal{L}_{\text{policy}} \approx \frac{1}{\text{N}} \sum_{s=1}^{\text{N}} \left( \log \pi^{k}(\vec{a}^s) - \log \pi^{k-1}(\vec{a}^s) \right) , 
\end{align}
where $\pi^{k}(\vec{a}^s)$ and $\pi^{k-1}(\vec{a}^s)$ are the probability density function (pdf) for sample $\vec{a}^s$ using GMM policy $\pi^{k}$ and $\pi^{k-1}$ respectively.
By combining all the modality-specific distillation loss functions, we have the following combined distillation loss
\begin{equation}
l_{\text{distill}}(\hat{s}_t, \hat{a}_t) = \lambda_i \mathcal{L}_{\text{image}} + \lambda_t \mathcal{L}_{\text{text}} + \lambda_e \mathcal{L}_{\text{extra}} + \lambda_p \mathcal{L}_{\text{policy}},
\end{equation}
where $\lambda_i, \lambda_t, \lambda_e, \text{and } \lambda_p$ are hyperparameters that control the balance between stability and plasticity of the policy throughout the learning process. %

\noindent \textbf{Final Optimization Objective.}
Putting all together, to update the policy, we optimize
\begin{align}
    \min_{\pi} J(\pi) &= \frac{1}{K} \sum_{k=1}^{K} 
    \mathbb{E}_{\substack{s_t, a_t \sim D^k \\ \hat{s}_t, \hat{a}_t \sim \hat{D}^k}} \left[ \sum_{t=0}^{L^k} 
    -\log \pi((a_t \cup \hat{a}_t) \mid \right. \notag \\
    &\left. \qquad (s \cup \hat{s})_{\leq t}; T^k) 
    + l_{\text{distill}}(\hat{s}_t, \hat{a}_t) \right].
\end{align}

Here, $D^k$ and $\hat{D}^k$ refer to the data distribution for the current task and memory exemplars, consisting of a subset of prior tasks' examples.

\section{Experimental Settings}
\label{sec:experiments}

\begin{table*}[th!]
\centering
\caption{Experimental results across three different LIBERO task suites. The reported values are averages from three seeds, including the mean and standard error. The best values are highlighted in bold, and the second-best values are underlined. The dash (-) indicates a failure to reproduce results. All metrics are measured based on success rates (\%).}
\label{tab:benchmark}
\resizebox{\textwidth}{!}{%
\begin{tabular}{lccccccccc}
\hline
\multirow{2}{*}{Method} & \multicolumn{3}{c|}{LIBERO-OBJECT}   & \multicolumn{3}{c|}{LIBERO-GOAL}    & \multicolumn{3}{c}{LIBERO-SPATIAL}\\ \cdashline{2-10}
        & FWT ($\uparrow$) & NBT ($\downarrow$) & \multicolumn{1}{c|}{AUC}~($\uparrow$) & FWT ($\uparrow$)       & NBT ($\downarrow$)       & \multicolumn{1}{c|}{AUC}~($\uparrow$) & FWT ($\uparrow$)       & NBT ($\downarrow$)       & AUC~($\uparrow$)      \\ \hline
Sequential               &   62.0 ($\pm$ 1.0)  &  63.0 ($\pm$ 2.0)   &  30.0 ($\pm$ 1.0)                       &      55.0 ($\pm$ 1.0)     &      70.0 ($\pm$ 1.0)     &     23.0 ($\pm$ 1.0)     &  72.0 ($\pm$ 1.0)   &  81.0 ($\pm$ 1.0)  &  20.0 ($\pm$ 1.0) \\
EWC~\cite{kirkpatrick2017overcoming}                       &   56.0 ($\pm$ 3.0)  &   69.0 ($\pm$ 2.0)  &            16.0 ($\pm$ 2.0)             &      32.0 ($\pm$ 2.0)     &      48.0 ($\pm$ 3.0)     &      6.0 ($\pm$ 1.0)        &  23.0 ($\pm$ 1.0)   &  33.0 ($\pm$ 1.0)  &  6.0 ($\pm$ 1.0) \\
ER~\cite{chaudhry2019tiny}                       &   56.0 ($\pm$ 1.0)  &   24.0 ($\pm$ 1.0)  &            49.0 ($\pm$ 1.0)             &      53.0 ($\pm$ 1.0)     &      36.0 ($\pm$ 1.0)     &      47.0 ($\pm$ 2.0)        &  65.0 ($\pm$ 3.0)   &  27.0 ($\pm$ 3.0)  &  56.0 ($\pm$ 1.0) \\

BUDS~\cite{zhu2022bottom}                        &   52.0 ($\pm$ 2.0)  &   21.0 ($\pm$ 1.0)  &            47.0 ($\pm$ 1.0)             &     50.0 ($\pm$ 1.0)      &      39.0 ($\pm$ 1.0)     &      42.0 ($\pm$ 1.0)        &   -  &  -  &  - \\
LOTUS~\cite{wan2024lotus}                        &   \underline{74.0} ($\pm$ 3.0)  &   \underline{11.0} ($\pm$ 1.0)  &            \underline{65.0} ($\pm$ 3.0)             &    \underline{61.0} ($\pm$ 3.0)       &      \underline{30.0} ($\pm$ 1.0)     &     \underline{56.0} ($\pm$ 1.0)        &   -  &  -  &  -  \\
\rowcolor{green!15} Ours                    &   \textbf{75.0} ($\pm$ 3.0)  &   \textbf{8.0} ($\pm$ 5.0)  &             \textbf{69.0} ($\pm$ 4.0)             &       \textbf{71.0} ($\pm$ 1.0)    &      \textbf{20.0} ($\pm$ 3.0)     &     \textbf{57.0} ($\pm$ 2.0)       &  \textbf{74.0} ($\pm$ 1.0)   &  \textbf{11.0} ($\pm$ 1.0)  &  \textbf{61.0} ($\pm$ 2.0)   \\ \hline
\end{tabular}%
}
\vspace{-.5em}
\end{table*}

\subsection{Training and Implementation Details}
We train our approach on a NVIDIA H100 GPU, and follow the data augmentation strategy proposed by \cite{liu2024libero}. For a fair comparison, our model shares the exact parameter configuration with the ResNet-T baseline and was trained with the same training hyperparameters. We train our model for $50$ epochs at every incremental step and we set the weight of our proposed regularization terms as follows; $\lambda_t$ and $\lambda_e$ are set to $0.05$ across all task suits. For LIBERO-OBJECT and LIBERO-SPATIAL, we use $0.05$ for both $\lambda_i$ and $\lambda_p$, whereas for LIBERO-GOAL, we increase their values to $0.25$. 
We evaluate our method against the following baselines:
\begin{itemize}
    \item \textbf{SEQUENTIAL}: This baseline involves naively fine-tuning new tasks sequentially using the ResNet-Transformer architecture from LIBERO~\cite{liu2024libero}.
    \item \textbf{EWC}~\cite{kirkpatrick2017overcoming}: A regularization based approach that regulates the network's weights by selectively updating relatively less important weights for prior tasks.
    \item \textbf{ER}~\cite{chaudhry2019tiny}: A rehearsal-based method that preserves a memory buffer containing samples from previous tasks and uses this buffer to facilitate the learning of new tasks. We impose a capacity limit of 1000 trajectories on the replay buffer.
    \item \textbf{BUDS}~\cite{zhu2022bottom}: A hierarchical policy baseline that utilizes multitask skill discovery. %
    \item \textbf{LOTUS} ~\cite{wan2024lotus}: A hierarchical imitation learning framework with experience replay that employs an open-vocabulary vision model for continual unsupervised skill discovery to identify and extract skills from unsegmented demonstrations. A meta-controller within LOTUS integrates these skills to manage vision-based manipulation tasks, allowing for effective LIL.
\end{itemize}

Results for the baseline methods are extracted from LIBERO~\cite{liu2024libero} and LOTUS~\cite{wan2024lotus}. However, we encountered difficulties reproducing the results for BUDS and LOTUS on LIBERO-SPATIAL, likely due to their challenges with skill discovery and spatial task generalization.

\subsection{Datasets}
\label{sec:datasets}

For our evaluations, we leverage a recently introduced lifelong imitation learning benchmark, LIBERO~\cite{liu2024libero}. LIBERO contains a diverse range of robotic tasks, features language-conditioned, diverse objects, sparse rewards, and long-horizon tasks.  
Our focus is on three specific suites: LIBERO-OBJECT, LIBERO-GOAL, and LIBERO-SPATIAL, each consisting of 10 tasks. These suites are crafted to explore the controlled transfer of knowledge regarding objects (declarative), task goals (procedural), and spatial information (declarative). In LIBERO-SPATIAL tasks, the robot is tasked with placing a bowl on a plate, distinguishing between two identical bowls that differ only in their spatial context. This requires ongoing learning and memorization of spatial relationships. In contrast, LIBERO-OBJECT tasks involve picking and placing distinct objects, which necessitates continual learning of different object types. Meanwhile, LIBERO-GOAL tasks use the same objects arranged spatially but differ in their goals, requiring the robot to learn new motions and behaviors.

\vspace{-2mm}
\subsection{Evaluation Metrics}
To evaluate how well policies perform in lifelong imitation learning for robot manipulation, we utilize three fundamental metrics: Forward Transfer (FWT), Negative Backward Transfer (NBT), and Area Under the Success Rate Curve (AUC), following~\cite{liu2024libero, wan2024lotus}. These metrics are based on success rates, providing a more dependable measure than training loss for manipulation tasks. FWT quantifies how effectively a policy adjusts to new tasks, with higher values signifying more effective learning and better integration of prior knowledge. NBT evaluates how well the policy preserves performance on earlier tasks while learning new ones, with lower values reflecting stronger retention of earlier performance. AUC provides a holistic measure of the policy's success across all tasks, with higher scores reflecting superior overall performance. %

\section{Results}
\label{sec:results}

In this section, we first evaluate whether our proposed distillation strategy aids the policy in leveraging existing skills while learning new manipulation tasks without forgetting in a lifelong imitation learning setup. %
Afterwards, we examine how our method’s performance changes at intermediate steps as the policy is trained on a sequence of manipulation tasks. Furthermore, we assess the effectiveness of our method in maintaining a consistent latent space. Finally, we present an ablation study to evaluate the contribution of each regularization term in our proposed multi-modal distillation-based LIL method.

\noindent
\textbf{Comparison to SOTA methods.} Table~\ref{tab:benchmark} provides a comprehensive evaluation of our proposed distillation-based LIL method, \coolname{}, in comparison to the current SOTA methods in LIBERO benchmark suites. We observe that regularization based strategy (\ie{} EWC\cite{kirkpatrick2017overcoming}) performs worse than the memory-replay based strategies across the task suites. The results indicate that our multi-modal distillation-based LIL strategy outperforms the baseline methods across all evaluation metrics in the LIBERO-OBJECT, LIBERO-GOAL, and LIBERO-SPATIAL task suites. In particular, our method achieves a $4\%$ higher AUC than LOTUS~\cite{wan2024lotus} in the LIBERO-OBJECT task suite, while showing similar FWT results. For the LIBERO-GOAL suite, our method achieves comparable AUC but shows a substantial improvement of $10\%$ in both the FWT and NBT metrics. Moreover, in the LIBERO-SPATIAL task suite, our approach exceeds ER\cite{chaudhry2019tiny} by approximately $15\%$ on the NBT metric and realizes a $5\%$ improvement in AUC. Overall, our proposed method exhibits robustness in leveraging previously acquired skills while also effectively learning new ones.

\begin{figure}[t!]
\hspace{-1em}
\centering
\includegraphics[width=.5\textwidth, scale=1]{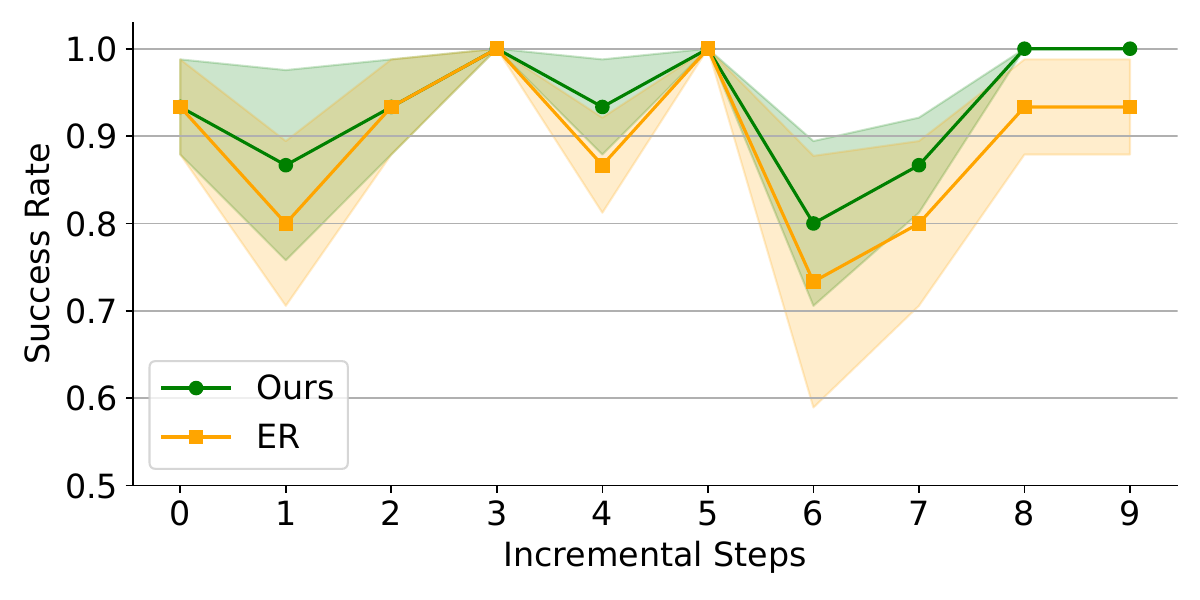}
\vspace{-1.5em}
\caption{%
The success rate across the incremental steps for ER and \coolname{} (Ours) on LIBEO-OBJECT task suite. Our method demonstrates a more consistent success rate across incremental steps compared to the ER baseline method. Higher values indicate better performance.
} 
\label{fig:success_rate_inc} 
\vspace{-1em}
\end{figure}

\noindent
\textbf{Performance Analysis.}
We assess the effectiveness of our proposed approach by measuring the success rate at each incremental step in the lifelong imitation learning scenario on the LIBERO-OBJECT task suite. For this evaluation, we compare our method to ER, and present the average success rate along with the standard error based on three seeds, as illustrated in Figure~\ref{fig:success_rate_inc}. The line plot shows that our method outperforms the Experience Replay (ER) baseline consistently. As training progresses, the performance gap between the two methods widens, highlighting the superior learning capacity of our proposed method. Furthermore, the lower standard error in our method suggests greater stability, clearly demonstrating its superior effectiveness in lifelong imitation learning tasks.

\begin{figure}[t]
\hspace{-1em}
\centering
\includegraphics[width=.5\textwidth, scale=1]{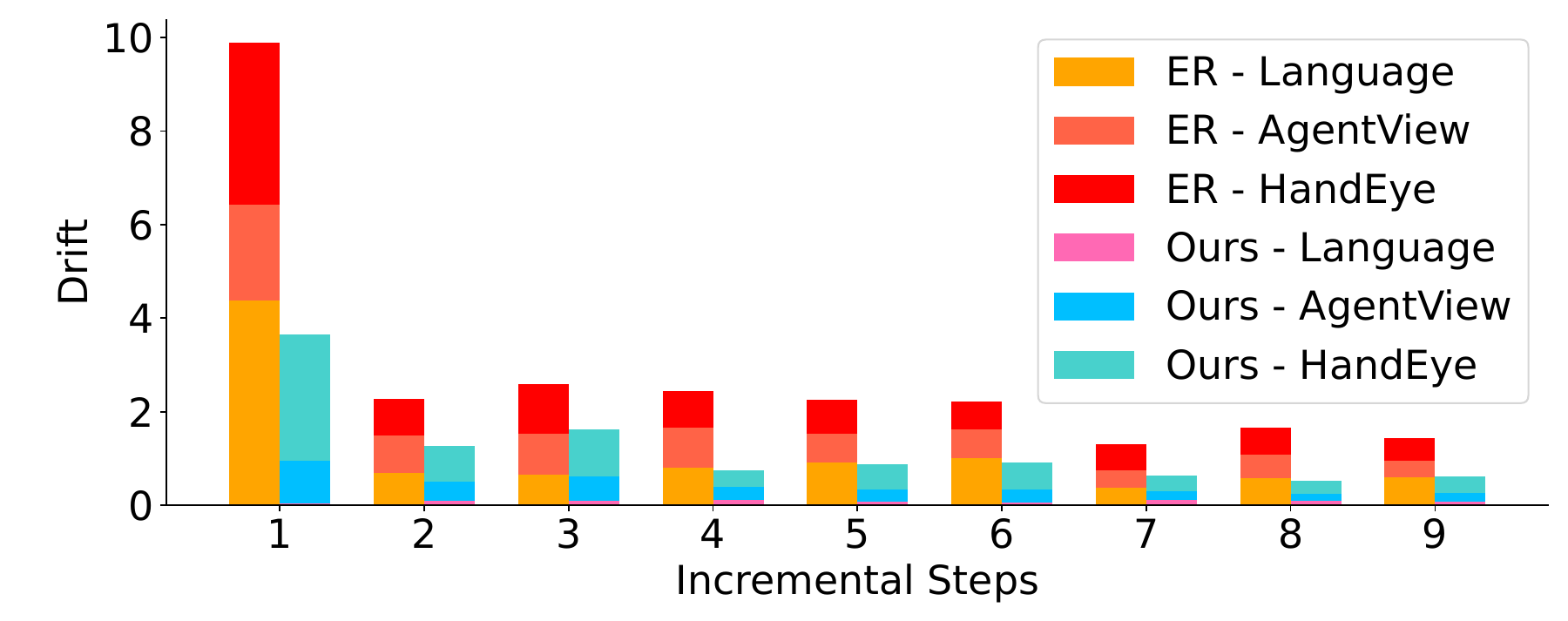}
\vspace{-1.5em}
\caption{Latent representation drift across the incremental steps for ER and \coolname{} (Ours) on LIBEO-OBJECT task suite. Drift is measured using the squared Euclidean distance between latent representations from policies at $t$ and $t-1$. Our method maintains a more consistent latent space across modalities compared to the ER baseline method. Lower values indicate better performance.
} 
\label{fig:latent_drift} 
\vspace{-1em}
\end{figure}

\noindent
\textbf{Latent Representation Drift Analysis.}
To assess the robustness of our proposed method in preserving a consistent latent space across different modalities during incremental steps in the lifelong imitation learning scenario on the LIBERO-OBJECT task suite, we compare our method against ER. We report the average drift in latent representations, calculated as the squared Euclidean distance between representations from the current and previous policies, averaged across three seeds, as illustrated in Figure~\ref{fig:latent_drift}. The bar plot illustrates that our method consistently exhibits less drift in latent representations across the Language, AgentView, and HandEye modalities compared to the Experience Replay (ER) baseline during incremental steps. The difference in performance is most noticeable in the later incremental steps, especially for Language and AgentView. This suggests that our method offers better stability in retaining learned skills over time.

\begin{table}[b!]
\centering
\caption{\label{tab:ablation_study} Ablation studies on the contribution of each component in our method. Experiments were performed on the LIBERO-OBJECT task suite using a seed value of 100.
}
\vspace{-0.9em}
\resizebox{\columnwidth}{!}{
\begin{tabular}{cccc|ccc}
\multicolumn{4}{c}{}                                                               & \multicolumn{3}{c}{}                                               \\ \hline
\multirow{2}{*}{$\text{L}_{\text{text}}$} & \multirow{2}{*}{$\text{L}_{\text{image}}$} & \multirow{2}{*}{$\text{L}_{\text{extra}}$} & \multirow{2}{*}{$\text{L}_{\text{action}}$} & \multicolumn{3}{c}{LIBERO-OBJECT} \\ 
&            &            &         & FWT $\uparrow$ & NBT $\downarrow$ & AUC $\uparrow$  \\ \hline
\cmark   &  \cmark   &   \cmark  &   \cmark  &   0.81    &   0.18    &   0.75    \\
\xmark   &  \cmark   &   \cmark  &   \cmark  &   0.81    &   0.20    &   0.68    \\
\cmark   &  \xmark   &   \cmark  &   \cmark  &   0.62    &   0.20    &   0.49    \\
\cmark   &  \cmark   &   \xmark  &   \cmark  &   0.70    &   0.22    &    0.55   \\
\cmark   &  \cmark   &   \cmark  &   \xmark  &   0.76    &   0.19    &   0.61    \\ 
\hline

\end{tabular} }
\end{table}

\noindent
\textbf{Ablation Studies.}
We conduct experiments on LIBERO-OBJECT using a seed value of 100 to examine the contribution of each distillation component in our strategy. The results (shown in Table~\ref{tab:ablation_study}) indicate that each regularization term is crucial for the consistent performance of the policy. Specifically, maintaining a consistent latent space for the vision modality is essential, as the performance significantly drops from $75\%$ to $49\%$ in AUC and $81\%$ to $62\%$ in FWT metric when distillation on the latent space for AgentView and HandEye views is not applied. Additionally, the impact of action distillation is notable; the AUC decreases by approximately $14\%$ without this regularization term. Furthermore, the absence of a regularizer for the extra modality results in a drop of about $10\%$ in FWT and $20\%$ in AUC. These findings highlight the importance of consistent latent representations of different modality information for preserving performance on prior tasks while learning novel manipulation tasks.

\section{Conclusion}
\label{sec:conclusion}

We propose a multi-modal distillation-based lifelong imitation learning approach for robot manipulation tasks. In this work, we focus on maintaining the latent space for different modalities and the action distribution throughout the learning experiences. To achieve this, we impose constraints on the alterations in the latent representations and action distributions between the prior and current policies. Specifically, we optimize the policy at step $T^k$ by minimizing the squared $L_2$ norm of latent features between the old and current encoders across different modalities, as well as the discrepancy between the prior and current GMM policies. Our proposed distillation strategy ensures a robust latent space alongside a GMM policy that preserves previously learned skills while adapting to new skills without forgetting. Through quantitative evaluation on the LIBERO task suites (\ie{} LIBERO-OBJECT, LIBERO-GOAL, and LIBERO-SPATIAL), we demonstrate that our proposed method significantly outperforms baseline methods across all evaluation metrics. For future work, we intend to investigate a memory-free distillation strategy for lifelong imitation learning that is robust to noise.

 \balance{}
\bibliographystyle{bibliography/IEEEtran}
\bibliography{ref}

\end{document}